\documentclass{isprs} 
\usepackage{natbib}
\usepackage{cuted}
\usepackage{multirow}
\usepackage{booktabs}
\usepackage{pifont}
\usepackage{subfigure}
\usepackage{float}
\usepackage{setspace}
\usepackage{geometry} 
\usepackage{epstopdf}
\usepackage{placeins}
\usepackage{amssymb}
\usepackage[labelsep=period]{caption}  
\usepackage[british]{babel} 
\usepackage[hang]{footmisc}
\usepackage{amsmath}
\usepackage{titlesec}
\usepackage{hyperref}
\hypersetup{
colorlinks=true, 
linkcolor=blue, 
citecolor=blue, 
urlcolor=blue 
}
\begin{document}

\title{SF-Recon: Simplification-Free Lightweight Building Reconstruction via 3D Gaussian Splatting}
\date{}



\author{
Zihan Li\thanks{Zihan Li and Tengfei Wang contributed equally to this work.},
Tengfei Wang\footnotemark[1],
Wentian Gan,
Hao Zhan,
Xin Wang\thanks{Corresponding author: Xin Wang, xwang@sgg.whu.edu.cn},
Zongqian Zhan
}

\address{School of Geodesy and Geomatics, Wuhan University, China PR.}



\abstract{Lightweight building surface models are crucial for digital city, navigation, and fast geospatial analytics, yet conventional multi-view geometry pipelines remain cumbersome and quality-sensitive due to their reliance on dense reconstruction, meshing, and subsequent simplification. This work presents SF-Recon, a method that directly reconstructs lightweight building surfaces from multi-view images without post-hoc mesh simplification. We first train an initial 3D Gaussian Splatting (3DGS) field to obtain a view-consistent representation. Building structure is then distilled by a normal-gradient-guided Gaussian optimization that selects primitives aligned with roof and wall boundaries, followed by multi-view edge-consistency pruning to enhance structural sharpness and suppress non-structural artifacts without external supervision. Finally, a multi-view depth-constrained Delaunay triangulation converts the structured Gaussian field into a lightweight, structurally faithful building mesh. Based on a proposed \textbf{\textit{SF}} dataset,  the experimental results demonstrate that our SF-Recon can directly reconstruct lightweight building models from multi-view imagery, achieving substantially fewer faces and vertices while maintaining computational efficiency.
}


\maketitle
\begin{strip}
  \centering
  \includegraphics[width=\textwidth]{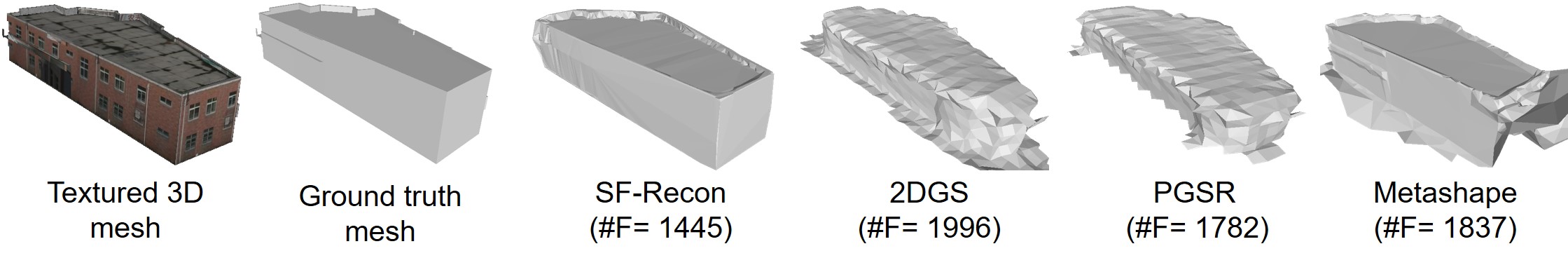}
  \captionof{figure}{Samples of Lightweight Reconstruction Methods, \#F denotes the number of triangular faces.}
  \label{fig:show}
\end{strip}

\section{Introduction}\label{Introduction}
Driven by advances in 3D vision and digital-twin technologies, the rapid generation of lightweight building surface models has become essential for indoor/outdoor navigation, geospatial analytics, and large-scale urban simulation \citep{LoD-Loc}. In practice, however, the prevailing pipeline for producing lightweight meshes from multi-view imagery remains lengthy and brittle. A standard photogrammetric workflow first reconstructs a sparse point cloud from structure-from-motion(SfM), then performs multi-view stereo (MVS) and meshing to obtain a detailed, high-fidelity surface model \citep{marching_cubes}, and finally applies mesh-simplification techniques, such as quadric error metrics (QEM) \citep{QEM}, to reduce the model complexity. This stage-wise process incurs substantial computational and engineering overhead and ties the quality of the lightweight mesh to the fidelity of the intermediate high-complexity mesh, undermining robustness and reproducibility in real deployments.

Neural scene representations, most notably 3D Gaussian Splatting (3DGS) \citep{3DGS}, offer a compelling alternative for multi-view reconstruction. Using explicit 3D Gaussian spheres as primitives, 3DGS performs end-to-end differentiable optimization for scene reconstruction, eliminating much of the algorithmic complexity inherent in conventional stage-wise workflows. Building on 3DGS, recent methods such as PGSR \citep{chen2024pgsr} and 2DGS \citep{2DGS} extract meshes from trained Gaussian fields via Truncated Signed Distance Function (TSDF) iso-surfacing \citep{TSDF}, with voxel size directly controlling the number of faces and vertices. Although these strategies can yield lightweight meshes by adjusting the voxel size to control face and vertex counts, they typically apply uniform meshing across regions, lack explicit modeling of building structure, and make limited use of the multi-view cues available during training. Consequently, salient building elements are often over-smoothed, and the reconstructed surfaces lack structural completeness and distinctiveness, as evident in the qualitative comparisons in Fig.~\ref{fig:show}.

To address these limitations, based on 3DGS, this paper proposes SF-Recon, a framework that reconstructs lightweight building surface meshes directly from multi-view images, eliminating the need for post-hoc mesh simplification. The pipeline consists of three main stages. First, an initial 3D Gaussian Splatting (3DGS) field is trained from input views, producing an explicit, view-consistent representation of scene appearance and geometry. Second, we introduce a tailored Gaussian Splatting for Building Framework, which isolates the subset of Gaussians corresponding to structural elements through a normal-gradient-guided optimization, and enhances structural sharpness while suppressing spurious primitives using a multi-view edge-consistency pruning strategy. Consequently, these can align the Gaussian field with building boundaries in an unsupervised manner, preserving key structures while maintaining compactness.
Finally, by leveraging the depth maps rendered during training, a multi-view depth-constrained Delaunay triangulation reconstructs a surface mesh that is both compact and structurally faithful, facilitating efficient downstream mapping and simulation.
In summary, our contributions are threefold:
\begin{enumerate}
\item To the best of our knowledge, we are the first to unleash 3DGS for directly lightweight building surface mesh reconstruction by taking only multi-view images as input, which can eliminate the need for post-hoc simplification from a pre-reconstructed complex mesh.
\item We introduce two key components: a normal-gradient-guided Gaussian optimization stage and a multi-view edge-consistency pruning strategy, which jointly guide the Gaussian field to explicitly encode building structures (e.g., rooflines and wall boundaries) in an unsupervised manner.
\item  A multi-view depth-constrained Delaunay meshing algorithm is developed, which converts the refined Gaussian field into a compact yet structurally consistent surface, enabling fast reconstruction while preserving building integrity.
\end{enumerate}

\section{Related Work}\label{Related Work}

Relevant works on 3DGS-based lightweight building surface reconstruction mainly follow two strands: (i) surface reconstruction using 3D Gaussian Splatting and (ii) strategies for lightweight mesh generation.

\subsection{Surface Reconstruction based 3D Gaussian splatting}\label{Related Work_1}
The advent of 3D Gaussian Splatting (3DGS) \citep{3DGS} extends neural rendering from radiance synthesis to explicit geometry recovery by modeling scenes as collections of parameterized Gaussian primitives. Early surface-oriented variants introduced geometric regularization to align Gaussians with the underlying surfaces. SuGaR \citep{sugar} couples Poisson-style meshing with constraints that pull Gaussians toward surface manifolds, achieving higher geometric accuracy than vanilla 3DGS. Nevertheless, the spatial irregularity of Gaussian distributions—particularly in low-texture or specular regions—can produce fragmented reconstructions.

To mitigate depth ambiguity and cross-view inconsistency, several works revise the underlying representation. 2DGS \citep{2DGS} projects 3D Gaussians onto view-consistent 2D disks, which stabilizes multi-view coherence and accelerates rendering and meshing, but can introduce depth bias due to the planar approximation. GOF \citep{GOF} instead operates on the opacity field induced by Gaussians and extracts geometry via level-set identification; although it improves completeness relative to 2DGS, enforcing global geometric consistency in complex scenes remains challenging.PGSR \citep{chen2024pgsr} employs planar-aligned Gaussians that encode surface normals and signed distances, enabling analytic projection and multi-view geometric regularization, and achieving strong geometric accuracy and photometric fidelity compared with contemporary baselines. In comparison, MeshSplat \citep{MeshSplat} leverages 2DGS as a bridge to synthesize novel-view images and transfers learned geometric priors to achieve sparse-view surface reconstruction with improved consistency and accuracy.

Although 3DGS-based methods can produce visually smooth, dense meshes, they are ill-suited to lightweight building reconstruction. Over-densification and irregular placement of Gaussian primitives inflate memory and introduce many redundant facets, yielding models that exceed lightweight budgets. For mesh extraction, TSDF iso-surfacing—used in methods such as 2DGS and PGSR—can throttle face counts via voxel size, but its uniform simplification degrades geometric fidelity and structural integrity, softening edges and distorting salient features (e.g., roof ridges and wall boundaries). Efficiently exploiting 3DGS to recover compact meshes that preserve complete building structure, thus remains a central open challenge.

\subsection{Lightweight Mesh Reconstruction}\label{Related Work_2}

Traditional pipelines for lightweight mesh reconstruction begin with high-precision dense point clouds, generate an initial detailed mesh via Poisson reconstruction or Delaunay triangulation, and then reduce complexity using quadric error metrics (QEM) \citep{QEM}, structural simplification \citep{structural_simplification}, or hierarchical strategies \citep{hierarchy_Strategy}.

Building-oriented lightweight reconstruction methods typically adopt low-polygonal modeling. \citet{Lafarg} identifies planar units via point-cloud structuring and assembles regular polygonal meshes. PolyFit \citep{polyfit} produces lightweight polygonal models by combining robust plane detection, patch-edge optimization, and topological reconstruction.  \citet{wang2023semantic} further introduces semantics to guide mesh simplification for semantic-aware lightweighting. WindPoly \citep{windpoly} leverages the winding number to achieve robust indoor–outdoor segmentation and topologically consistent reconstruction. However, these approaches are highly data-dependent: reconstruction quality is tightly coupled to the fidelity of the input point cloud or mesh. Moreover, their pipelines are lengthy and serial, leading to error accumulation and reduced stability. These limitations motivate a shift toward neural representations and deep learning that recover lightweight, structurally faithful building models directly from readily acquired images.

Based on 3D Gaussian Splatting (3DGS), several multi-view methods—EdgeGaussians \citep{edgegaussians}, SketchSplat \citep{ying2025sketchsplat}, and CurveGaussian \citep{CurveGaussian}, recover 3D edge contours directly from images. While effective for wireframe extraction, these approaches do not produce lightweight surface meshes and typically depend on auxiliary pre-trained networks to generate edge maps used as supervision. To the best of our knowledge, an end-to-end method that reconstructs structurally complete and geometrically faithful lightweight meshes directly from multi-view images (without external supervision or pre-trained models) remains scarce. To fill this gap, this work proposes SF-Recon, which employ 3DGS and ingests only multi-view building imagery without any edge priors for directly yielding lightweight building surface meshes that preserve salient geometric structure.

\section{Preliminaries of 3D Gaussian Splatting}\label{sec:3dgs_prelim}

3D Gaussian Splatting (3DGS) represents a scene as a set of parameterized 3D Gaussians. Each Gaussian is defined by a center $\mu\in\mathbb{R}^3$ and a covariance matrix $\Sigma\in\mathbb{R}^{3\times 3}$, which controls its spatial extent and orientation. The Gaussians are defined as:
\begin{equation}
G(x)=\exp\!\left(-\tfrac{1}{2}(x-\mu)^{\top}\Sigma^{-1}(x-\mu)\right)
\end{equation}
Gaussians are typically initialized from a sparse SfM point set \citep{sfm} and optimized to a dense, view-consistent scene representation. Each primitive carries an opacity $\alpha\in[0,1]$ and a color attribute $c$ parameterized by spherical-harmonic coefficients, enabling photorealistic view synthesis. All parameters are differentiable and learned by minimizing a photometric objective over training views, commonly a mixture of photometric loss $L_1$ and structural similarity loss $L_{SSIM}$:
\begin{equation}
{L} = \lambda {L}_1 + (1-\lambda) {L}_{SSIM}
\end{equation}
To render a novel image, 3D Gaussians are projected to a view-dependent 2D elliptical splat and perform front-to-back alpha compositing. For a ray $r$, with splats $\{i\}$ sorted by depth, the color is estimated as:
\begin{equation}
c(r)=\sum_{i\in\mathcal{M}} T_{i-1}\,\alpha_i\,c_i,\qquad
T_{i-1}=\prod_{j< i}(1-\alpha_j)
\end{equation}
where $\mathcal{M}$ denotes the set of contributing splats, $\alpha_i$ their opacities, and $T_{i-1}$ the accumulated transmittance.

A key property of 3DGS is that Gaussian positions, shapes, opacities, and colors are continuously refined during training. In this work, we impose structure-aware constraints and multi-view pruning on the Gaussian primitives to bias the field toward building edges (e.g., roof ridges and wall boundaries) and suppress non-structural clutter, yielding a representation better suited for lightweight building surface reconstruction.
\begin{figure*}[htbp]
    \centering
    \includegraphics[width=1\linewidth]{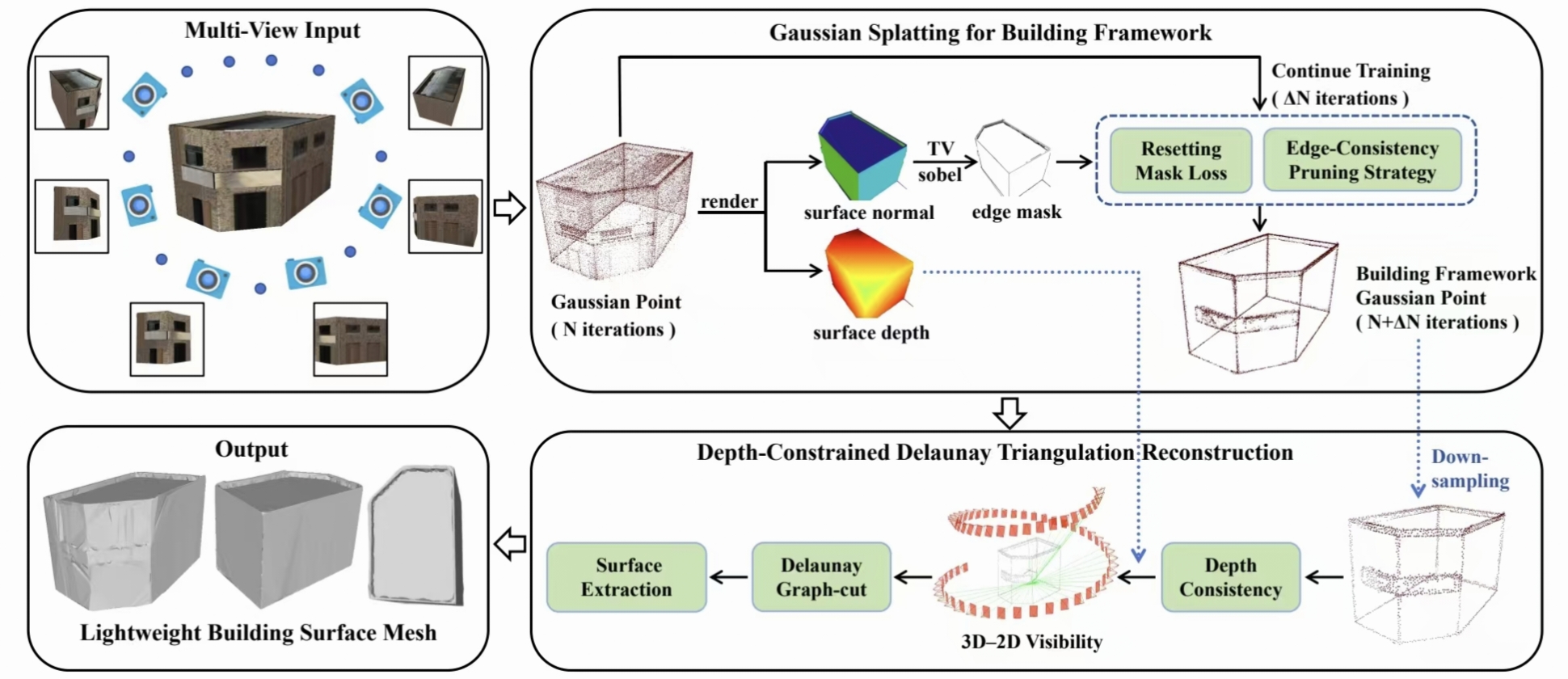}
    \caption{The pipeline of SF-Recon}
    \label{fig:pipeline}
\end{figure*}

\section{Method} \label{sec:Method}
To overcome the complexity and quality limitations of conventional lightweight building reconstruction, based on 3DGS, we propose \textbf{SF-Recon}, which recovers lightweight building surfaces directly from multi-view images without post-hoc mesh simplification. Our method first optimizes a 3D Gaussian Splatting (3DGS) field to emphasize the building framework via a normal-gradient-guided Gaussian optimization and a multi-view edge-consistency pruning strategy (Section~\ref{sec:Method1}). Leveraging depths rendered during training, it then performs a multi-view, depth-constrained Delaunay triangulation to produce a compact yet structurally consistent surface mesh (Section~\ref{sec:Method2}). The overall pipeline is shown in Fig.\ref{fig:pipeline}

\subsection{Gaussian Splatting for Building Framework} \label{sec:Method1}
Gaussian Splatting for Building Framework builds on the 2DGS training framework and introduces a normal-gradient-guided Gaussian optimization and a multi-view edge-consistency pruning strategy during training to refine the Gaussian field, thereby improving the accuracy of the building-framework representation.

\subsubsection{Normal-Gradient-Guided Gaussian Optimization}
To ensure that the Gaussian field accurately captures the building framework, edge-localization guidance is injected into the optimization process. This guidance biases Gaussian primitives to concentrate along structural boundaries during split/merge updates while suppressing occupancy in non-edge regions. Instead of relying on pre-trained single-view edge detectors, surface normals are estimated in situ during training. These cues are inherently multi-view consistent and do not require external supervision. Concretely, the primitives are temporarily frozen to compute stable normal-gradient signals, which are then incorporated into a structure-aware loss. This loss promotes alignment with high-gradient regions and penalizes non-edge coverage. The resulting gradients reallocate density toward edges and reduce clutter, yielding a Gaussian field that explicitly encodes the building framework.

\textbf{Edge-Mask Extraction from Frozen Gaussian Field.} The Gaussian Field of the 20k iteration is frozen. Per-view normal maps are then rendered from the current Gaussian field, which provides sufficiently accurate normals at this stage. To obtain edge cues, each normal map is first denoised using total-variation (TV) denoising, followed by the computation of image gradients using Sobel filters. Pixels whose gradient magnitude exceeds a threshold of 0.5 are labeled as edges, resulting in a binary edge mask for each view.

The total-variation (TV) denoising problem seeks a denoised image \(u\) by minimizing
\begin{align}
E(u; I)=\min_{u}\int_{\Omega}\!\Big(\,\lvert\nabla u(x)\rvert^{2}+\lambda\,\lvert u(x)-I(x)\rvert^{2}\Big)\,dx
\end{align}
where \(\nabla u\) denotes the spatial gradient and \(\lambda>0\) balances smoothness and data fidelity.

The per-pixel gradient magnitude is then computed as
\begin{align}
\lvert\nabla I(i,j)\rvert=\sqrt{G_x(i,j)^2+G_y(i,j)^2}
\end{align}
where \(G_x(i,j)\) and \(G_y(i,j)\) are the horizontal and vertical gradient components at pixel \((i,j)\).

Edges are obtained by thresholding the gradient magnitude with a parameter \(T>0\):
\begin{align}
M(i,j)=
\begin{cases}
1, & \text{if } \lvert\nabla I(i,j)\rvert>T\\
0, & \text{otherwise}
\end{cases}
\end{align}
with \(M:\Omega\to\{0,1\}\) the resulting binary edge mask.

\textbf{Loss Function Regarding Edge Mask.} After extracting building-edge masks, densification of the Gaussian field is resumed for the subsequent 10k iteration optimization, and the objective is redefined to incorporate a per-view binary edge mask. This adjustment biases the Gaussian primitives toward structural boundaries while down-weighting non-edge regions. Let $m(i)\in\{0,1\}$ denote the edge mask and $\varepsilon>0$ a small constant for numerical stability. The mask-averaged $\ell_1$ loss is
\begin{align}
L_{\text{1\_refined}}
= \frac{\sum_{i=1}^{N} m(i)\,\bigl|I_{\mathrm{render}}(i)-I_{\mathrm{gt}}(i)\bigr|}
       {\sum_{i=1}^{N} m(i)+\varepsilon}
\end{align}
The SSIM loss is defined analogously:
\begin{align}
L_{\text{SSIM\_refined}}
= \frac{\sum_{i=1}^{N} m(i)\,\bigl[1-\mathrm{SSIM}\!\bigl(I_{\mathrm{render}}(i),\,I_{\mathrm{gt}}(i)\bigr)\bigr]}
       {\sum_{i=1}^{N} m(i)+\varepsilon}
\end{align}
To further enforce normal alignment, the depth-derived normal supervision used in 2DGS\citep{2DGS} is also applied, and a mask-weighted normal loss is computed:
\begin{align}
L_{\text{normal\_refined}}
= \frac{\sum_{i=1}^{N} m(i)\,\bigl[1-\mathrm{clip}\!\left(\mathbf{n}(i)\!\cdot\!\mathbf{n}_{\mathrm{depth}}(i),-1,1\right)\bigr]}
       {\sum_{i=1}^{N} m(i)+\varepsilon}
\end{align}
where $\mathbf{n}(i)$ and $\mathbf{n}_{\mathrm{depth}}(i)$ are unit normals from the renderer and the depth map, respectively.
The total loss combines the image and normal terms:
\begin{align}
L_{\text{total}} &= \lambda_1 \cdot L_{\text{1\_refined}}+ \lambda_2 \cdot L_{\text{ssim\_refined}} \nonumber \\
&\quad + \lambda_3 \cdot L_{\text{normal\_refined}}
\end{align}

\subsubsection{Multi-View Edge-Consistency Pruning Strategy}

In the subsequent 10k training iterations, misassigned Gaussian primitives can degrade geometric fidelity and inflate model complexity. To address this, a multi-view edge-consistency pruning strategy is introduced, which systematically removes primitives that provide little or no support for structural boundaries. Each Gaussian primitive $G_i$ receives an \emph{edge score} $e_i\in[0,1]$ that indicates how consistently it projects onto building edges across various views; higher scores reveal stronger support for edge (roof ridges and wall–facade intersections).

Let $\mathcal{C}$ be the set of training images. For image $j\in\mathcal{C}$ with projection matrix $\mathbf{P}_j$ and per-view edge mask $\mathcal{E}_j\subset\Omega_j$, the projection of the 3D location $\mathbf{x}_i=(x_i,y_i,z_i)^{\top}$ is
\begin{align}
\pi_j(\mathbf{x}_i)=\Pi\!\big(\mathbf{P}_j\,\tilde{\mathbf{x}}_i\big),\qquad
\tilde{\mathbf{x}}_i=\begin{pmatrix}\mathbf{x}_i\\[2pt]1\end{pmatrix}
\end{align}
where $\Pi(\cdot)$ denotes homogeneous normalization to image coordinates. The per-view edge-visibility indicator is defined as:
\begin{align}
v_{ij} = \mathbb{I}\left( \pi_j(\mathbf{x}_i) \in \mathcal{E}_j \right) \cdot \mathbb{I}\left( \mathbb{V}_{ij} \right)
\end{align}
where \( \mathbb{V}_{ij} \) is the visibility of \( G_i \) in view \( j \), which is determined by the renderer’s visibility check.
\begin{align}
e_i=\frac{1}{|\mathcal{C}|}\sum_{j\in\mathcal{C}} v_{ij}
\end{align}
Primitives with $e_i$ below a threshold $\tau$ are pruned:
\begin{align}
\mathcal{G}_{\mathrm{pruned}}=\{\,G_i\in\mathcal{G}\mid e_i<\tau\,\}
\end{align}
This pruning is applied at regular intervals over the subsequent 10K training iterations. Iteratively removing low-score primitives produces a sparser, boundary-aligned Gaussian field that concentrates capacity on salient edges, improving structural fidelity while reducing redundancy for downstream meshing.

\subsection{Depth-Constrained Delaunay Triangulation Reconstruction} \label{sec:Method2}

This section consolidates the meshing pipeline: Starting from building-framework points, reliable 3D point cloud to 2D image visibility correspondences are established under multi-view depth constraints. The surface is subsequently extracted via a visibility-driven Delaunay graph cut, yielding a watertight and structurally regular mesh. The pipeline is robust to view-dependent sampling-density variations and occlusions, and offers tunable parameters to balance geometric detail and surface smoothness.

\subsubsection{Depth-Constrained 3D--2D Visibility Validation}

Establishing a stable one-to-one association between each 3D point and its valid 2D observations is challenging because point clouds produced by Gaussian training are typically dense, exhibit view-dependent visibility, and yield overlapping or occluded projections on the image plane. To mitigate these factors, we construct a robust 3D--2D visibility relation using the multi-view depth maps produced during training as a constraint. For each 3D point \(p\) and candidate view \(i\), we project \(p\) to the image; if the projection lies within the image bounds, we sample the depth map at that location via bilinear interpolation to obtain \(d_{\text{img}, ip}\). We then compute the expected depth \(d_{\text{exp}, ip}\) from the camera model along the viewing ray and accept the correspondence when the ray-level depth consistency condition holds:
\begin{align}
\label{eq:ray-depth-consistency}
|d_{\text{exp}, ip} - d_{\text{img}, ip}| \leq \epsilon_{\text{abs}} + \epsilon_{\text{rel}} d_{\text{exp}, ip}
\end{align}
where \(\epsilon_{\mathrm{abs}}\) accounts for depth-independent effects such as interpolation error and minor calibration offsets, while \(\epsilon_{\mathrm{rel}}\) models depth errors that grow in proportion to scale. A point is deemed visible and not occluded in view \(i\) only when inequality~\eqref{eq:ray-depth-consistency} holds. Recording these validated pairs yields a stable, accurate, and reproducible correspondence set that supports subsequent visibility-driven surface reconstruction.

\subsubsection{Visibility-Driven Delaunay Graph-Cut for Surface Extraction}

Following \citet{labatut_robust}, we perform visibility-driven surface reconstruction within the Delaunay tetrahedralization of the input building-outline point cloud. The tetrahedralization defines a set of tetrahedra and their shared triangular facets. Surface extraction is formulated as an $s$--$t$ minimum-cut problem on the dual graph, with edge capacities that combine a visibility likelihood term and a geometric regularization term:
\begin{align}
J(\Sigma) \;=\; \Phi_{\mathrm{vis}}(\Sigma) \;+\; \beta\,\Phi_{\mathrm{geo}}(\Sigma)
\end{align}
where $\Sigma$ is the extracted surface, $\Phi_{\mathrm{vis}}$ is the visibility cost, $\Phi_{\mathrm{geo}}$ is the geometric quality cost, and $\beta$ balances the two terms.

The visibility cost measures how well each facet is supported by the observations. Facets that lie closer to the reprojected samples and are traversed by more consistent viewing rays receive a lower penalty ad are thus favored by the cut. For a facet $f$,define the visibility cost:
\begin{align}
\Phi_{\mathrm{vis}}(f) \;=\; \alpha_{\mathrm{vis}}\!\left( 1 - e^{-\,d^{2}/(2\sigma^{2})} \right)
\end{align}
where $\alpha_{\mathrm{vis}}$ is a scaling factor related to per-facet visibility , $d$ denotes the distance from the reprojected 3D sample to the candidate facet $f$ measured along the viewing ray, and $\sigma$ controls the decay of the visibility kernel.

The geometric cost discourages irregular or degenerate facets, preventing jagged cuts inside elongated or skewed tetrahedra. For a facet $f$, define the geometric cost:
\begin{align}
\Phi_{\mathrm{geo}}(f) \;=\; 1 - \min\!\bigl(\cos\varphi,\; \cos\psi\bigr)
\end{align}
where $\varphi$ and $\psi$ are the angles between the segment joining the circumcenters of two adjacent tetrahedra and the normal to their shared facet.

By minimizing the objective function \(J(\Sigma)\), each tetrahedron is assigned an inside or outside label. The final building surface is then formed by the interface facets shared by adjacent outside/inside tetrahedra:
\begin{equation}
\Sigma_{\mathrm{building}}
\;=\;
\bigcup_{F_{\mathrm{shared}}} f_{\mathrm{shared}},
f_{\mathrm{shared}} \;=\; T^{\mathrm{out}}_{k}\,\cap\, T^{\mathrm{in}}_{k}
\end{equation}
where \(\Sigma_{\mathrm{building}}\) denotes the extracted building surface, \(T^{\mathrm{out}}_{k}\) and \(T^{\mathrm{in}}_{k}\) are tetrahedra labeled as outside and inside, respectively, and \(f_{\mathrm{shared}}\) is their common triangular face.

After surface extraction, we apply a post-filtering step based on a maximum edge-length criterion to remove spurious or isolated triangles, producing a smoother surface. The proposed meshing pipeline preserves geometric regularity and smoothness while allowing the mesh detail to adapt to the local point-cloud density, enabling multi-level reconstruction.

\subsection{ Experimental Datasets}

\begin{figure}[htbp]
    \includegraphics[width=1\linewidth]{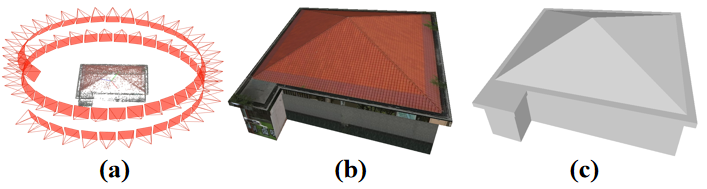}
    \caption{The sample of \textbf{\textit{SF}} dataset. (a) Camera trajectory around the building, (b) Textured 3D model of the building, (c) Ground truth mesh of the building.}
    \label{fig:dataset}
\end{figure}

Owing to the absence of multi-view datasets that include both real textures and ground-truth lightweight meshes for buildings, we introduce a novel dataset, referred to as \textbf{\textit{SF}}, consisting of 10 manually reconstructed buildings, each with photorealistic, baked textures. For each building, 80 views are rendered at a resolution of \(1000\times1000\), using a spiral trajectory that encircles the building while varying elevation. This ensures comprehensive multi-view coverage and sufficient overlap between views. Each image is generated with calibrated camera intrinsics and extrinsics, facilitating reproducible multi-view geometry experiments. Some samples are presented in Fig.\ref{fig:dataset}.

\section{Experiments}
\begin{figure*}[htbp]
  \centering
  \includegraphics[width=\textwidth]{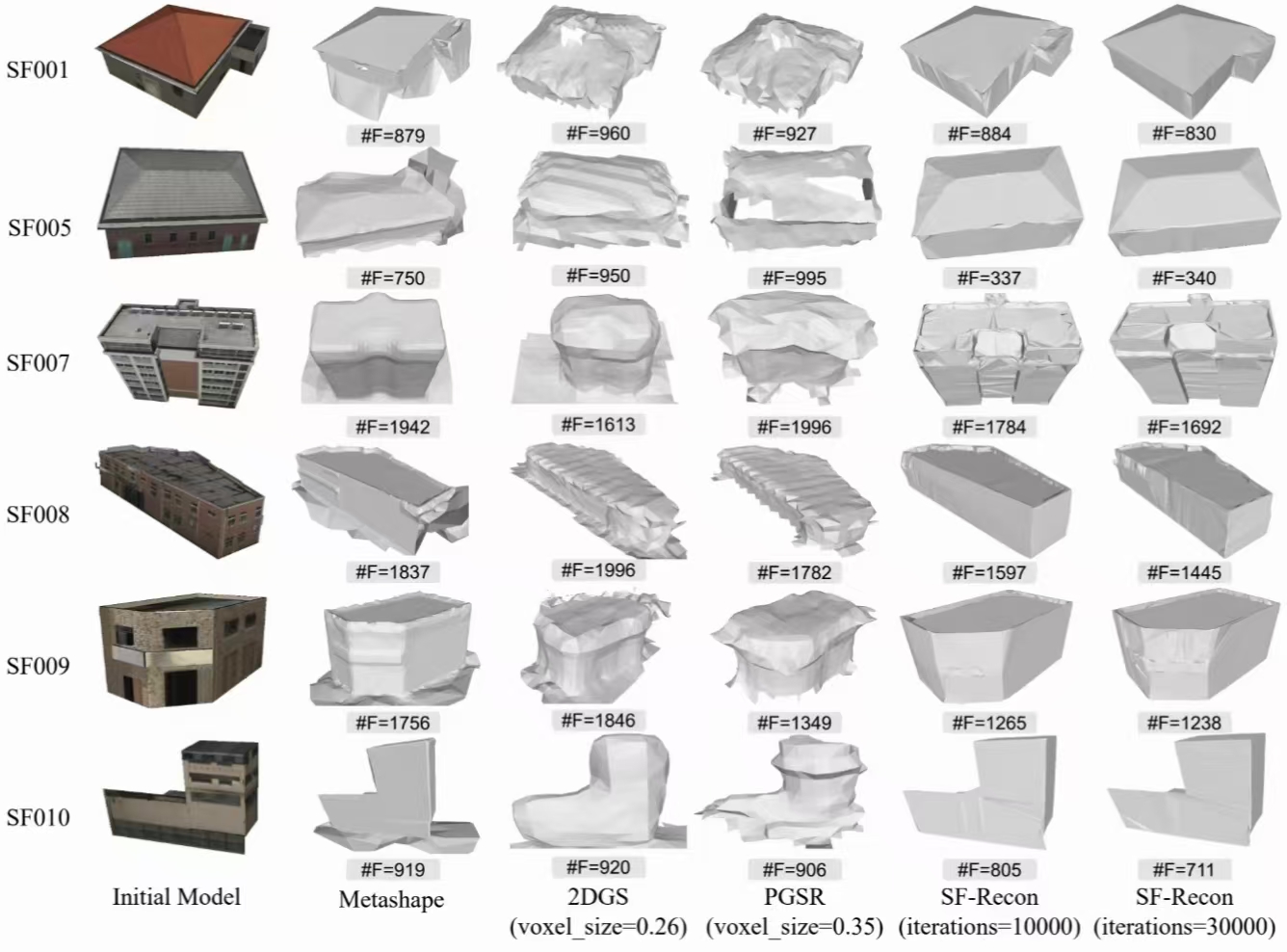}
  \caption{Qualitative comparison on \textbf{\textit{SF}} dataset. \#F denotes the number of triangular faces.}
  \label{fig:evaluation}
\end{figure*}
\begin{figure*}[htbp]
  \centering
  \includegraphics[width=\textwidth]{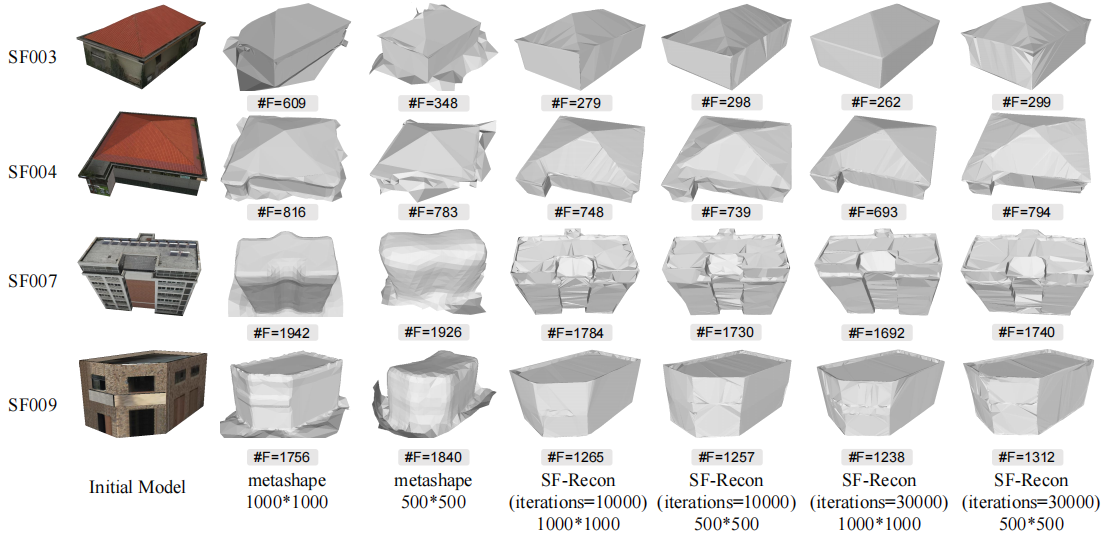}
  \caption{Qualitative comparison under different resolutions on \textbf{\textit{SF}} dataset.}
  \label{fig:resolution}
\end{figure*}

\textbf{Baselines and Implementation.}
SF-Recon is compared with three baselines: PGSR, 2DGS, and the commercial-software Metashape \citep{Metashape}, as no existing approach is specifically designed for the direct reconstruction of lightweight building meshes from multi-view images, these methods are the most relevant for our comparison. All experiments are conducted on a workstation configured with an Intel Xeon Gold 6133 CPU and an NVIDIA RTX 4090 GPU. To ensure a fair comparison, in PGSR and 2DGS the Gaussian fields are trained on multi-view images for 30,000 iterations, after which TSDF fusion is applied to the predicted depth maps to extract meshes. The number of generated mesh elements is controlled by adjusting the TSDF voxel size. For Metashape, we adjust the reconstruction settings to produce meshes with reduced polygon counts, so that the reconstructions from all methods contain approximately the same number of faces and vertices.To evaluate the impact of the number of training iterations on mesh quality, we cap training at 10,000 iterations, scale all stages proportionally, and report a second set of reconstructions.

\textbf{Metrics.}
Following \citep{wang2023semantic}, to evaluate the performance of various methods, the number of faces and vertices, total reconstruction time, and mesh reconstruction error are reported. The mesh reconstruction error is defined as the root mean square error (RMSE) of the distances from each vertex of the reconstructed mesh to its nearest triangle on the ground-truth mesh.
\begin{align}
\mathrm{RMSE}
=
\sqrt{
\frac{
\displaystyle \sum_{v \in m_{\mathrm{rec}}^{B}}
\displaystyle \min_{\Delta \in M_{\mathrm{gt}}}
\left\| v - P_{\Delta}(v) \right\|^{2}
}{
\left| m_{\mathrm{rec}}^{B} \right|
}
}
\end{align}

\begin{table*}[htbp]
\centering
\caption{Quantitative evaluation results on the \textbf{\textit{SF}} datasets. Best values are in \textbf{bold}.}
\label{tab:sf-quant}
\renewcommand{\arraystretch}{1.5}
\begin{tabular}{lcccc}
\toprule
\textbf{Method} & \textbf{Faces Count} & \textbf{Vertices Count} & \textbf{Time (mins)} & \textbf{RMSE (m)} \\
\midrule
PGSR                  & 1063 & 563 & 28.3 & 0.2874 \\
2DGS                  & 1137 & 561 & 12.3 & 0.3809 \\
Metashape             & 1153 & 518 & \textbf{2.7} & 0.2652 \\
SF-Recon(10000)         & 926 & 472 & 2.9 & 0.0763 \\
\textbf{SF-Recon(30000)} & \textbf{914} & \textbf{467} & 9.5 & \textbf{0.0456} \\
\bottomrule
\end{tabular}
\end{table*}

\newcommand{\cmark}{\ding{51}} 
\newcommand{\xmark}{\ding{55}} 

\begin{table*}[htbp]
\centering
\caption{Ablation study of SF-Recon modules. \textbf{NG-GO}: normal-gradient-guided optimization; \textbf{ECP}: multi-view edge-consistency pruning; \textbf{DCD}: depth-constrained Delaunay triangulation. Best values are in \textbf{bold}.}
\label{tab:ablation}
\renewcommand{\arraystretch}{1.25}
\setlength{\tabcolsep}{6.5pt}
\begin{tabular}{lccccccc}
\toprule
\multirow{2}{*}{\textbf{Configuration}} & \multicolumn{3}{c}{\textbf{Modules}} & \multicolumn{4}{c}{\textbf{Quantitative results}} \\
\cmidrule(lr){2-4} \cmidrule(lr){5-8}
& \textbf{NG-GO} & \textbf{ECP} & \textbf{DCD} & \textbf{Faces} & \textbf{Vertices} & \textbf{Time (min)} & \textbf{RMSE (m)} \\
\midrule
Ablation 1               & \cmark & \xmark & \cmark & 1016 & 527 & 9.9 & 0.1002 \\
Ablation 2               & \cmark & \cmark & \xmark & 1266 & 544 & 9.6 & 0.4781 \\
Ablation 3               & \xmark & \xmark & \cmark & 1001 & 505 & 12.1 & 0.1962 \\
\textbf{Full (SF-Recon)} & \cmark & \cmark & \cmark & \textbf{914} & \textbf{467} & \textbf{9.5} & \textbf{0.0456} \\
\bottomrule
\end{tabular}
\end{table*}

where \(m_{\mathrm{rec}}^{B}\) denotes the set of vertices of the reconstructed building mesh, and
\(M_{\mathrm{gt}}\) is the ground-truth model represented as a triangular mesh.
For a triangle \(\Delta \in M_{\mathrm{gt}}\), \(P_{\Delta}(v)\in\Delta\) is the closest point (orthogonal
projection) of vertex \(v\) onto the surface of \(\Delta\).

\subsection{Comparisons with Other Methods}

Qualitative experiments are conducted on the proposed \textbf{\textit{SF}} datasets, which are shown in Fig.~\ref{fig:evaluation}. It can be seen that Metashape successfully generates high-quality meshes but with numerous redundant facets and irregular surfaces, even under simplified settings. In contrast, 2DGS and PGSR produce smoother yet overly simplified meshes, as voxel-grid iso-surfacing lacks explicit structural priors (e.g., roof ridges, wall boundaries) and does not fully leverage multi-view geometric consistency during training, resulting in the loss of important building details. Additionally, PGSR’s multi-view consistency loss is sensitive to repetitive textures commonly found in urban scenes. SF-Recon strikes a favorable balance between structural fidelity and compactness, producing lighter meshes while preserving key building structures, such as the roofs and walls.

Quantitative results in Table~\ref{tab:sf-quant} substantiate the advantage of SF-Recon. With comparable number of face and vertex, SF-Recon achieves a markedly lower RMSE than other competing methods. Even with only 10,000 training iterations, SF-Recon maintains competitive accuracy and achieves a runtime comparable to that of Metashape.

\subsection{Ablation Studies}

To evaluate the effectiveness of each module in SF-Recon, three ablations are designed, as summarized in Table~\ref{tab:ablation}. Each ablation replaces or removes specific modules to investigate their individual impact on the reconstruction compactness, structural fidelity and runtime,with all ablation conducted under 30,000 iterations.

\textbf{Ablation 1.}
This variant retains normal-gradient-guided Gaussian optimization and depth-constrained Delaunay triangulation but omits the multi-view edge-consistency pruning.
Without this pruning step, spurious Gaussian primitives from non-edge regions are preserved, leading to redundant structures and noticeable surface noise, which degrade model compactness.

\textbf{Ablation 2.}
Here, normal-gradient-guided Gaussian optimization and edge-consistency pruning are applied, but the final mesh is generated via TSDF fusion instead of multi-view depth–constrained Delaunay triangulation. Although edge pruning suppresses clutter and improves sparsity, TSDF-based methods fail to construct triangulated meshes between different building boundaries, resulting in the reconstruction of only the building boundaries. Therefore, the TSDF method is not suitable for lightweight mesh reconstruction when combined with SF-Recon.

\textbf{Ablation 3.}
This configuration preserves only depth-constrained triangulation. Without gradient guidance, Gaussian primitives fail to align with structural boundaries, leading to blurred roof ridges, softened wall intersections, and a general loss of geometric sharpness, despite maintaining a similar mesh size.

Overall, these ablation studies confirm that normal-gradient-guided optimization, multi-view edge-consistency pruning, and depth-constrained Delaunay triangulation are all essential.
Together, they enable SF-Recon to achieve lightweight yet structurally faithful building-surface reconstructions.

\subsection{Discussion}
In addition, we find that the resolution of input images can severely influence the final lightweight surface model; thus, a corresponding discussion is presented. In particular, SF-Recon, 2DGS, PGSR, and Metashape are evaluated on images with resolutions of \(1000 \times 1000\) and \(500 \times 500\). As shown in Table~\ref{tab:sfres} and Fig.\ref{fig:resolution}, SF-Recon performs consistently well at both resolutions and is more insensitive to resolution changes. In contrast, Metashape works better at higher resolutions and is significantly degraded at lower resolutions, occasionally failing to reconstruct entire buildings. 2DGS and PGSR, however, remain worse across both resolutions. These findings can imply that traditional geometric reconstruction methods, like Metashape, are highly resolution-dependent, requiring higher-resolution imagery to recover fine details. On the other hand, 3DGS-based methods such as SF-Recon maintain stable reconstruction quality even with lower-resolution inputs, demonstrating superior robustness for lightweight building modeling.

\begin{table*}[htbp]
\centering
\caption{Quantitative evaluation on the \textbf{\textit{SF}} dataset at two resolutions: \(\mathbf{1000\times1000}\) vs \(\mathbf{500\times500}\). Best values are in \textbf{bold}.}
\label{tab:sfres}
\renewcommand{\arraystretch}{1.35}
\setlength{\tabcolsep}{6pt}
\begin{tabular}{lcccccccc}
\toprule
\multirow{2}{*}{\textbf{Method}} &
\multicolumn{4}{c}{\(\mathbf{1000\times1000}\)} &
\multicolumn{4}{c}{\(\mathbf{500\times500}\)} \\
\cmidrule(lr){2-5} \cmidrule(lr){6-9}
 & \textbf{Faces} & \textbf{Vertices} & \textbf{Time (min)} & \textbf{RMSE (m)}
 & \textbf{Faces} & \textbf{Vertices} & \textbf{Time (min)} & \textbf{RMSE (m)} \\
\midrule
PGSR        & 1063 & 563 & 28.3 & 0.2874 & 1057 & 585 & 20.3 & 0.2887 \\
2DGS        & 1137 & 561 & 12.3 & 0.3809 & 1168 & 566 & 10.2 & 0.3819 \\
Metashape   & 1153 & 518 & \textbf{2.7} & 0.2652 & 1137 & \textbf{495} & \textbf{1.5} & 0.2841 \\
SF-Recon (10000)   & 926 & 472 & 2.9 & 0.0763 & 994 & 506 & 2.1 & 0.1092 \\
\textbf{SF-Recon (30000)} & \textbf{914} & \textbf{467} & 9.5 & \textbf{0.0456} &
\textbf{986} & 503 & 7.2 & \textbf{0.0842} \\
\bottomrule
\end{tabular}
\end{table*}    

Despite better high-quality lightweight models can be achieved across various resolutions, our \textbf{\textit{SF-Recon}} has several limitations. Compared to traditional labor-intensive pipelines that simplify high-complexity meshes, SF-Recon retains an excess of vertices along edges, yielding redundant faces and an insufficiently simplified mesh. Additionally, boundary-mask extraction from normals is effective when multi-view imagery provides high coverage and rich textures, but its performance deteriorates in texture-poor regions. Moreover, the computational efficiency requires further improvement.

\section{Conclusion}
In this work, we introduced SF-Recon, a novel framework using 3DGS to reconstruct lightweight building surfaces directly from multi-view images, eliminating the need for post-hoc mesh simplification. By leveraging 3D Gaussian Splatting (3DGS) combined with a normal-gradient-guided optimization and multi-view edge-consistency pruning, SF-Recon captures building structure with high fidelity while maintaining computational efficiency. The approach preserves essential building features such as rooflines and wall boundaries, achieving higher accuracy with a similar number of faces and vertices compared to other methods. Additionally, SF-Recon demonstrates superior robustness to variations in image resolution, performing consistently well across different resolutions, unlike traditional methods that are highly resolution-dependent. SF-Recon shows promising results but has yet limitations, including higher face and vertex counts, lower accuracy, and performance degradation in texture-poor regions. Future work will focus on improving efficiency and robustness.

{
    \begin{spacing}{1.10}
        \normalsize
        \bibliography{ISPRSguidelines_authors} 
    \end{spacing}
}

\end{document}